\documentclass{article}

\PassOptionsToPackage{numbers, compress}{natbib}
\usepackage[preprint]{neurips_2019}
\usepackage[utf8]{inputenc} % allow utf-8 input
\usepackage[T1]{fontenc}    % use 8-bit T1 fonts
\usepackage{hyperref}       % hyperlinks
\usepackage{url}            % simple URL typesetting
\usepackage{booktabs}       % professional-quality tables
\usepackage{amsfonts}       % blackboard math symbols
\usepackage{nicefrac}       % compact symbols for 1/2, etc.
\usepackage{microtype}      % microtypography

\usepackage{amsmath}
\usepackage{amssymb}
\usepackage{url}

\usepackage{multirow} % mutli row tables
\usepackage{caption} % required for subfigure environment
\usepackage{subcaption} % required for subfigure environment

\usepackage{graphicx}

\usepackage[T1]{fontenc}    % use 8-bit T1 fonts
\usepackage[utf8]{inputenc} % allow utf-8 input
\usepackage{nicefrac}       % compact symbols for 1/2, etc.

\usepackage{amsthm}
\usepackage{amssymb} % assumes amsmath package installed
\usepackage{mathtools}
\usepackage{mathrsfs}

\usepackage{graphics} % for pdf, bitmapped graphics files
\usepackage{tikz}
\usetikzlibrary{shapes, shapes.geometric, arrows, calc}
\usetikzlibrary{bayesnet}
\usepackage{pgfkeys}
\usepackage{standalone}

\newcommand{\realnumbers}{\mathbb{R}}

\newcommand{\expected}{\mathbb{E}}
\newcommand{\expectation}{\mathbb{E}}

\newcommand{\loss}{\mathcal{L}}

\newcommand{\eigval}{\Lambda}
\newcommand{\eigvec}{V}
\newcommand{\convolve}{\ast}

\usepackage{xspace}
\newcommand{\data}{x}

\newcommand{\dataset}{\mathcal{D}}

\newtheorem{theorem}{Theorem}
\newtheorem{lemma}[theorem]{Lemma}

\newtheorem{definition}{Definition}
\newtheorem{corollary}{Corollary}

\newtheorem{example}{Example}

\title{
Quantifying the effect of representations \\
on task complexity
}

\author{Julian Zilly \and Lorenz Hetzel \and Andrea Censi \and Emilio Frazzoli% <-this % stops a space
\thanks{The authors are with the Institute for Dynamic Systems and Control, ETH Zurich, Zurich, Switzerland. Correspondence to \texttt{\{jzilly, hetzell\}@ethz.ch}}%
}

\begin{document}

\maketitle
\thispagestyle{empty}
\pagestyle{empty}

\begin{abstract}
We examine the influence of input data representations on learning complexity.
For learning, we posit that each model implicitly uses a candidate model distribution for unexplained variations in the data, its noise model.
If the model distribution is not well aligned to the true distribution, then even relevant variations will be treated as noise. 
Crucially however, the alignment of model and true distribution can be changed, albeit implicitly, by changing data representations. 
``Better'' representations can better align the model to the true distribution, making it easier to approximate the input-output relationship in the data without discarding useful data variations.
To quantify this alignment effect of data representations on the difficulty of a learning task, we make use of an existing task complexity score and show its connection to the representation-dependent information coding length of the input.
Empirically we extract the necessary statistics from a linear regression approximation and show that these are sufficient to predict relative learning performance outcomes of different data representations and neural network types obtained when utilizing an extensive neural network architecture search. 
We conclude that to ensure better learning outcomes, representations may need to be tailored to both task and model to align with the implicit distribution of model and task.
\end{abstract}

\section{Introduction}

\begin{figure}[ht]
    \centering
    \includegraphics[width=0.25\columnwidth]{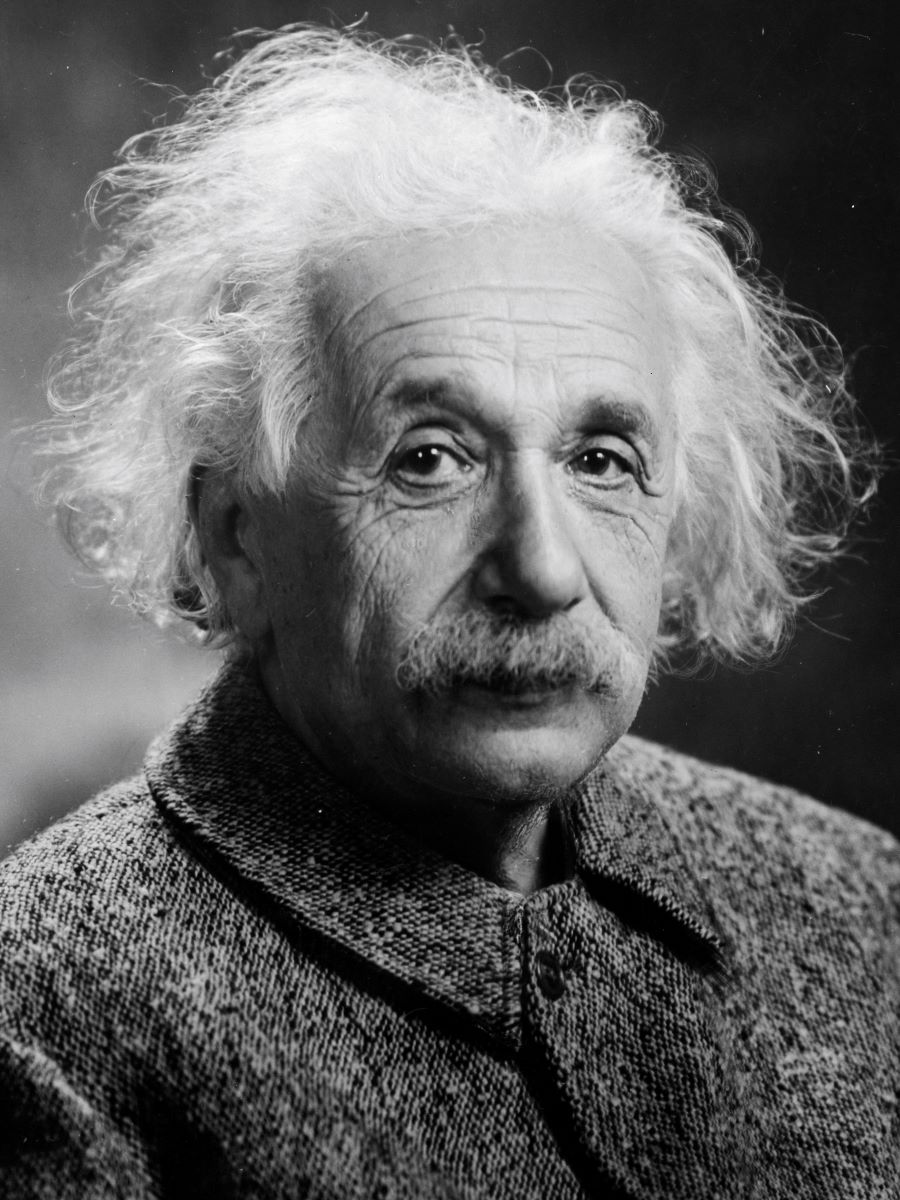}
    \includegraphics[width=0.25\columnwidth]{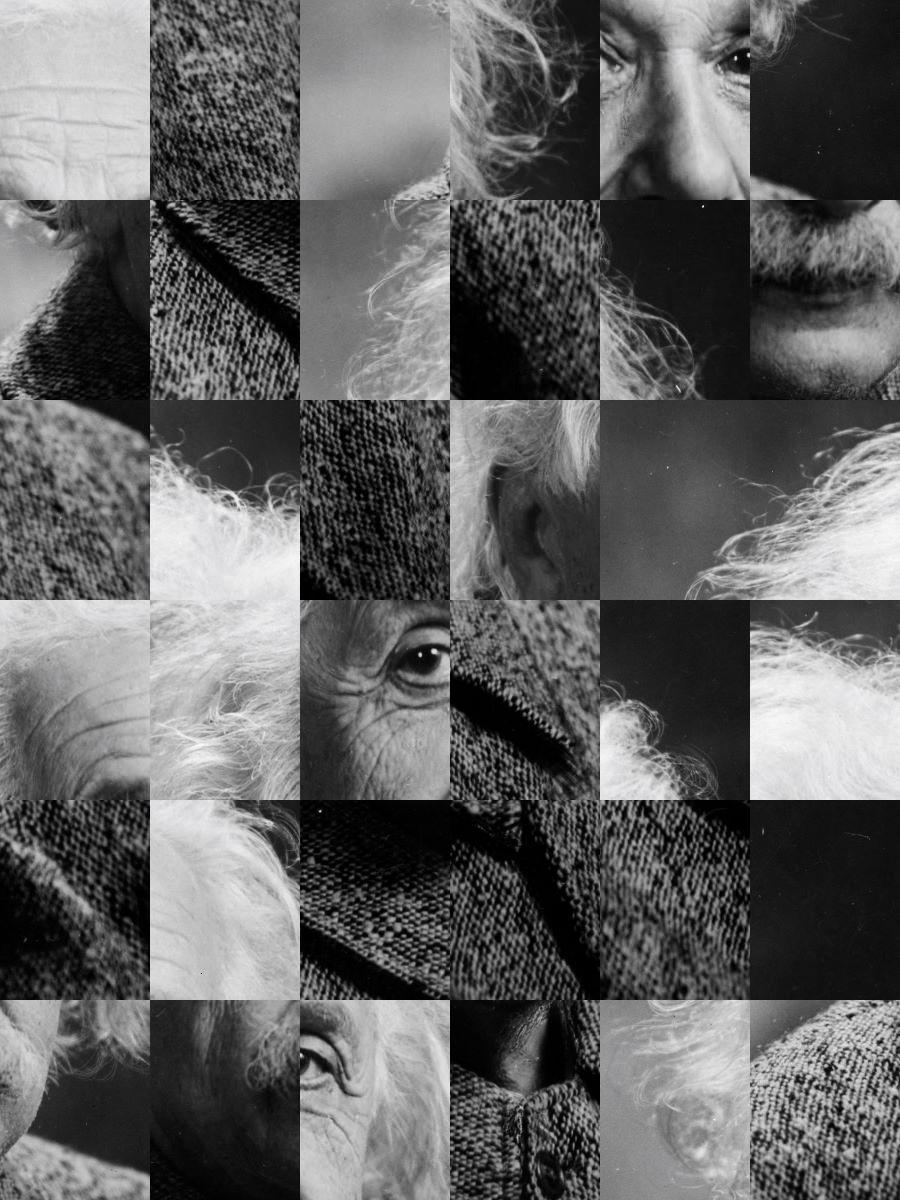}
    \caption{These images contain different representations of the same information. However, one of the two is much easier for us to understand. We posit that, for our nervous system, one of the two images has a lower \emph{expected coding length} for the task of recognizing the person.}
    \label{fig:overview_introduction}
\end{figure}

Sometimes perspective is everything. 
While the information content of encoded data may not change when the way it is represented changes, its usefulness can vary dramatically (see Fig.~\ref{fig:overview_introduction}). 
A ``useful'' representation then is one that makes it easy to extract information of interest. 
This in turn very much depends on who or which algorithm is extracting the information. 
Evidently the way data is encoded and how a model ``decodes'' the information needs to match. 

Historically, people have invented a large variety of ``data representations'' to convey information. 
An instance of this theme is the heliocentric vs. geocentric view of the solar system. 
Before the heliocentric viewpoint was widely accepted, scholars had already worked out the movements of the planets~\citep{theodossiou2002pythagoreans}. 
The main contribution of the new perspective was that now the planetary trajectories were simple ellipses instead of more complicated movements involving loops\footnote{For a clear illustration, see for example\\ \url{http://astronomy.nmsu.edu/geas/lectures/lecture11/slide01.html}}.

In a machine learning context, many have experimented with finding good data representations for specific tasks such as speech recognition~\citep{logan2000mel}, face recognition~\citep{hsu2002face}, for increased robustness in face detection~\citep{podilchuk1998face}, and many others. 
Yet no clear understanding has emerged of why a given representation is more suited to one task but less for another. 
We cast the problem of choosing the data representation for learning as one of determining the ease of encoding the relationship between input and output which depends both on how the data is represented and which model is supposed to encode it.

\emph{Contribution:}
In this work, we argue that learning a task is about encoding the relationship between input and output. 
Each model implicitly has a way of encoding information, where some variations in the data are easier to encode than others. 
Armed with this insight, we empirically evaluate different data representations and record what impact data representations have on learning outcomes and types of networks found by automated network optimization. 
Most interestingly, we are able to show that relative learning outcomes of neural architecture searches for different representations can be predicted by measuring the variance of weights of linear regression to these problems, where the feature extraction of linear regression is adapted depending on whether fully connected or convolutional networks are approximated.

\subsection{Related Work}

This work aims to bring us a bit closer to understanding the effect of data representations on what makes a given learning task easier or harder. 

\textbf{Data representations:} 
Data representations have been optimized for a long time. In fact there is a rich theory of linear invertible representations for both finite and infinite dimensional spaces called \emph{Frame Theory}~\citep{christensen2016introduction}. 
Specific popular examples of frames are Wavelets~\citep{mallat1999wavelet} and Curvelets~\citep{candes2000curvelets}. 
Empirically tested only on Imagenet, Uber research~\citep{gueguen2018faster} showed that using a data representation closer to how JPEG encodes information may help to create faster residual network architecture with slightly better performance. 
In a similar spirit in a robotics context, Grassmann and Kahrs~\citep{Grassmann2019OnTM} evaluated learning performance on approximating robot dynamics using various common robot dynamics data representations such as Euler angles. 
What is more common in deep learning is to adapt the network architecture to the task at hand. 
An intriguing recent example taking this idea a step further are Weight Agnostic Neural Networks~\citep{gaier2019weight} which have been designed to already ``function'' on a task even when starting from randomly initialized weights. 

\textbf{Measuring learning difficulty:} % Literature related to conditioning
Already in the ninenties, \citet{thornton1995measuring} posed the question of how to measure how easy or difficult a learning task is and related the difficulty to information theoretic measures called the information gain (mutual information) and the information gain ratio 
introduced in the context of decision trees by \citet{Quinlan1986, quinlan2014c4}.  
\citet{ho2002complexity} take a different road by comparing several possible scores to assess the difficulty of classification learning problems such as linear separability and feature efficiency. 
More commonly, instead of judging task difficulty, there is a vast literature on feature selection~\citep{guyon2003introduction}, e.g. judging how suitable a feature is for a given learning problem. 
Desirable features are reliably selected for a learning task~\citep{meinshausen2010stability} and ideally are highly predictive of the output variable. 
More recently based on PAC-Bayes bound~\cite{mcallester2003pac} considerations, Achille~\cite{information_in_deep_nets} derived an expression for learning task complexity based on the expected training performance and the KL-divergence from posterior and prior of the weights. Likewise there is a large literature on Kolmogorov complexity that tries to encode the algorithmic complexity of an object with applications such as measuring \emph{data complexity}~\cite{li2006data}.

\section{Data Representations and Task Complexity}

The objective of learning can be phrased as finding a function that minimizes the uncertainty of the output given the input while discarding as much task irrelevant information as possible. 
In information theoretic language, this viewpoint was introduced by \citet{tishby2000information} and extended by \citet{achille2018emergence} in the form of the objective of the Information Bottleneck (IB) Lagrangian.
Given an input $x$, a model encoding $z$, an output $y$ and mutual information $I(\cdot;\cdot)$, the IB-Lagrangian~\cite{tishby2000information} aims to minimize the following cost function:
\begin{align*}
    \mathcal{L}(p(z|x)) = I(x;z) - \beta I(y;z)
\end{align*}
The model is supposed to find an encoding $z$ of data $x$ that maximizes the mutual information to output $y$ while also minimizing the mutual information with $x$.
We consider the influence data representations have on optimizing the above objective and begin by defining what we mean when we talk about a \emph{data representation}. 
\begin{definition}
A data representation $r \in \mathcal{X}$ is the output of an invertible mapping $m(\cdot): \mathcal{X} \to \mathcal{X}$ applied to the ``original'' data $x \in \mathcal{X}$. 
\begin{align*}
    r = m(x), \forall x \in \mathcal{X}, \quad x = m^{-1}(r), \forall r \in \mathcal{X}
\end{align*}
\end{definition}

Therefore, all data representations are in a sense ``equivalent''. Since they are invertible they share the same information content $I(x;y) =I(m(x);y)$. 
Yet clearly how data is represented does influence learning. 
As a ``worst case'', an encrypted version of a dataset is unlikely to be useful for a learning task.
A representation $r=m(x)$ induces a new candidate distribution.
As an example, assume the task of fitting $y=x^2 + \mathcal{N}(0,\sigma^2)$ for $x\in [0, 1]$ via linear regression. Then the representations $r_1=x, r_2=x^2=m(r_1)$ are invertible to each other on this interval, yet the latter allows for a much better fit using linear regression than the former. In the former case, unexplained variations are assumed to be noise. Both candidate distribution $q_r(y|x)$ and $q_r(x)$ can change with representation $r$.

To understand what impact a data representation may have we will employ the idea of expected coding length $\expectation[l(x)]$ and focus on what happens when we choose the ``wrong code''. 
From Information Theory~\citep{cover2012elements}, we learn that the most efficient encoding we can possibly find is lower bounded by the entropy of the distribution we are trying to compress. 
In this case, we assume that we have a candidate distribution $q(x)$ that we are trying to fit to the true distribution $p(x)$. 
The expected coding length of our candidate distribution can then never be smaller than the entropy of the true distribution~\citep{cover2012elements}: $\expectation[l(x)] = H_q(x) \geq H_p(x)$.

\begin{theorem}[(Wrong code) Theorem 5.4.3~\citep{cover2012elements}]
The expected length $l(x)= \lceil \log \frac{1}{q(x)} \rceil$ under $p(x)$ of the code assignment satisfies
\begin{align}\label{eq:wrong_code}
    H(p) + D_{KL}(p||q) \leq \expectation_p [l(X)] =: \hat{H}_q(x) < H(p) + D_{KL}(p||q) + 1,
\end{align}
where $D_{KL}(p||q)$ is the Kullback-Leibler divergence between $p$ and $q$.
\end{theorem}

Critical to the point we are making, we will assume that any function family $\mathcal{F}$ has an associated candidate distribution $q(x)$ with its own ``codes'' $l(x) = \lceil \log \frac{1}{q(x)} \rceil$ and expected coding length $\hat{H}_q(x)$ through which it measures how uncertain a variable is, e.g. linear regression usually assumes a normal distribution for all variations that are not explained linearly. 
The difficulty with assuming a candidate distribution is that available data may not follow the same distribution. 
Given such a mismatch, the model will overestimate the entropy of the distribution as shown in theorem~\ref{eq:wrong_code}. 
\begin{lemma}[Representation-Model-Alignment]\label{lemma:repr_alignment}
Assuming candidate distributions $q_{r_1}(x), q_{r_2}(x)$ and representations $r_1, r_2$ with $D_{KL}(p(x)||q_{r_1}(x)) > D_{KL}(p(x)||q_{r_2}(x)) + 1$ we have that 
\begin{align*}
    \hat{H}_{q_{r_1}}(x) > \hat{H}_{q_{r_2}}(x)
\end{align*}
\begin{proof}
From theorem~\ref{eq:wrong_code} we know that
$ H(p(x)) + D_{KL}(p(x)||q(x))\leq \hat{H}_q(x)) < H(p(x)) + D_{KL}(p(x)||q(x)) + 1$. Thus using the ineq. relation $\hat{H}_{q_{r_1}}(x) \geq A, \hat{H}_{q_{r_2}}(x) < B \Rightarrow \hat{H}_{q_{r_1}}(x) - \hat{H}_{q_{r_2}}(x) > A-B$ we get  $\hat{H}_{q_{r_1}}(x) - \hat{H}_{q_{r_2}}(x) > H(p(x)) + D_{KL}(p(x)||q_{r_1}(x)) - H(p(x)) - D_{KL}(p(x)||q_{r_2}(x)) - 1 = D_{KL}(p(x)||q_{r_1}(x)) - D_{KL}(p(x)||q_{r_2}(x))) - 1 > 0 $
\end{proof}
\end{lemma}

Critically for real-world situations, the wrong code theorem invalidates the assumption that the estimated entropy does not change when an invertible transformation is applied. 
Both entropy and mutual information do indeed not change, yet in practice one does not necessarily have access to the true distribution $p(x)$ but only $q(x)$. 
The closer representation $r$ aligns the model candidate distribution to the true distribution, the smaller the data coding length $\hat{H}_{q_{r}}(x)$ will be. 
In this sense there are thus better and worse invertible data representations given a model and learning task.

Coming back to the IB-Lagrangian objective, an attractive option to evaluate the influence of data representations on how difficult a learning task is via the following task complexity defined by Achille~\cite{information_in_deep_nets}.
\begin{align}\label{eq:complexity}
    C_\beta (\mathcal{D}, P, Q) = \expectation_{w \sim Q(w|\mathcal{D})}[\loss_{\mathcal{D}}(p_w(y|x))] + \beta D_{KL}(Q(w|\mathcal{D})||P(w))
\end{align}

This term is closely aligned to the IB-Lagrangian defined above and trades-off the performance of the learned model on the \emph{training set} and the distance of the weights of this model to the prior weights as expressed by the KL-divergence between the two distributions. 
The second term here has the crucial function of quantifying the confidence that can be placed in any loss values computed on the training data similar to PAC-Bayes bounds~\cite{mcallester2003pac}. 
The further the posterior is from the prior belief, the more data is necessary to place confidence in the obtained predictions. 
As a metaphor of why the uncertainty estimate matters imagine the task of predicting the birthrate in Mongolia from the movements of the stock market for a given month. 
Most certainly one will be able to correlate the two. This is a common occurrence called \emph{spurious correlation}~\citep{fan2012variance}. % making us fools of randomness~\citep{taleb2005fooled}. 

Furthermore, we learn from Achille~\cite{information_in_deep_nets} that the KL-term is, in expectation over the dataset distribution $\mathcal{D}$ equal to the mutual information between weights $w$ and dataset distribution $\dataset$: $\expectation_{\mathcal{D}}[D_{KL}(Q(w|\mathcal{D})||P(w))] = I(w;\mathcal{D})$. Additionally, by assuming that the posterior weights $w_i$ independently follow a log-normal distribution $\log \mathcal{N}(-\alpha_i/2, \alpha_i)$ as found in~\cite{achille2018emergence} we can write the information in the weights $w_i$ as $I(w;\mathcal{D}) = - \frac{1}{2} \sum_{i=1}^{\text{dim}(w)} \log \alpha_i + C$ with $\alpha_i = \log(1+\text{var}(w_i))$.

We are interested in finding a connection between different data representations and their effect on the task complexity defined in Eq.~\ref{eq:complexity} above. 
By definition the task complexity is centered around changes from prior to posterior weights, yet we will see that albeit initially implicitly, there is a connection to the way data is represented.
For the following calculations we consider the particular regime which is relevant for several applications of interest, such as object detection.

\begin{definition}[Distillation  regime]
In the \emph{distillation regime} we assume that:
\begin{enumerate}
    \setlength{\itemsep}{0pt}
    \item The samples $x$ have very high entropy $H(x)$. Thus we assume $\text{var}(w_{ij}) \ll 1$ holds for linear regression. 
    \item The entropy of $y$ is small with respect to the entropy of $x$.
\end{enumerate}
\end{definition}
\begin{example}
Typical object detection tasks are in the distillation regime. The entropy of images is high (property 1), while labels are compactly represented (property 2). 
\end{example} 

In the following, we consider the case of linear regression as an ``approximation'' to the more complicated neural networks. 
This is motivated by empirical observations~\cite{goodfellow2014explaining} that neural networks, on a spectrum from linear to highly nonlinear, are likely to be close in behavior to linear models in many ways. In particular, when using ReLU-activations~\cite{relu}, the neural network is locally linear in the units that are not affected by the activation function. 

\begin{lemma}\label{lemma:complexity}
We define $x \in \mathbb{R}^n, y \in \mathbb{R}^m, w \in \mathbb{R}^{m \times n}$ and dataset distribution $\mathcal{D}=p(x,y)$.
We consider the linear regression setting in the distillation regime with $y_j = \sum_{i}w_{ji}x_{i} + \epsilon, \epsilon \sim \mathcal{N}(0, \sigma_\epsilon I_{m \times m})$ and assume in the distillation regime that the variance of residuals is equal to the noise variance $\sigma_r = \sigma_\epsilon$. The trained weights are assumed to each independently follow the posterior distribution $w_{ij} = \delta_{ij} \cdot \hat{w}_{ij}$ with log-normal candidate distribution $\delta_{ij} \sim \log \mathcal{N}(-\alpha_{ij}/2, \alpha_{ij})$ and deterministic $\hat{w}_{ij}$.
Then the following lower bound holds. 
\begin{align*}
    I(w;\mathcal{D}) \geq \frac{m}{n} \hat{H}_{q_\mathcal{N}}(x) + \text{const.} \geq \frac{m}{n} (H(x) + D_{KL}(p||q_\mathcal{N}))  + \text{const.}
\end{align*}
\begin{proof}
\begin{align*}
    I(w; \mathcal{D}) &= -\frac{1}{2} \sum_{i,j}^{\text{dim}(w)} \log(\alpha_{ij}) + C = -\frac{1}{2} \sum_{i,j}^{\text{dim}(w)} \log(\log(1 + \text{var}(w_{ij})) + C \\
           &\overset{\text{1.}}{\approx} -\frac{1}{2} \sum_{i,j}^{\text{dim}(w)} \log(\text{var}(w_{ij})) + C \overset{\text{lin. reg.}}{=} -\frac{1}{2} \sum_{i,j}^{\text{dim}(w)} \log(\sigma^2_{\epsilon_j} (X^TX)^{-1}_{ii}) + C \\
        &\overset{\text{Jensen's ineq.}}{\geq} -\frac{1}{2} \sum_{j}^{m} \log \Big(\sum_i^n (X^TX)^{-1}_{ii} \Big) + n\log(\sigma_{\epsilon_j}) + C \\
        &= -\frac{1}{2} \sum_{j}^{m} \log \Big(\text{tr}(\Sigma_x^{-1}) \Big) + n\log(\sigma_{\epsilon_j}) + C \\
           &\overset{\text{tr}(\Sigma) \geq n \det(\Sigma)^{1/n}}{\geq} -\frac{1}{2} \sum_{j}^{m} \log(n\det(\Sigma_x)^{-1/n}) + C_2, \text{ collecting const. terms}\\  
           &\overset{H_{\mathcal{N}}(x)=\frac{1}{2}\log(\det(2\pi e \Sigma_x))}{=} \frac{m}{n} \hat{H}_{q\mathcal{N}}(x) + C_3 \geq \frac{m}{n} (H(x) + D_{KL}(p||q_\mathcal{N})) + C_4,
\end{align*}
where we are assuming that the covariance matrix $\Sigma_x$ is symmetric and positive semi-definite. 
\end{proof}
\end{lemma}
We note that the computed entropy $\hat{H}_q(x)$ is based on the linear regression assumptions of a linear functional relationship and Gaussian noise for $q(x)$. Since $p(x)$ does not in general follow this assumed distribution $q(x)$ this leads to the expected coding length $\hat{H}_q(x)=H_{\mathcal{N}}(x)$ and not to the true entropy $H(x)$. 
The closer the true distribution is to a Gaussian, the better this approximation is and the smaller the $D_{KL}(p||q)$ correction term is. 
From the task complexity definition (Eq.~\ref{eq:complexity}) we derive the following task complexity score heuristic. 
If a given threshold of performance is achieved, solutions are favored which have lower mutual information of weights to dataset distribution. 

\begin{definition}[TCS]
Given random variables $x,y, w$, mutual information $I(w;\mathcal{D})$, threshold $t$, and loss on training data $\expectation_w[\loss_{\mathcal{D}}(p_w(y|x))]$, the \emph{task complexity score} (\emph{TCS}) is defined as
\begin{align*}
    \emph{TCS}(\mathcal{D}, w, t) := \begin{cases} 
    \frac{1}{I(w;\mathcal{D})}, &\text{ if } \quad \expectation_w[\loss_{\mathcal{D}}(p_w(y|x))] < t \\
    0, &\text{ otherwise}
    \end{cases}
\end{align*}
\end{definition}

\begin{corollary}[Representation effect on task complexity]
The smaller the KL-divergence $D_{KL}(p||q_r)$ of candidate distribution $q_r(x)$ of a representation $r=m(x)$ is, the smaller the lower bound in lemma~\ref{lemma:repr_alignment} and the larger the \emph{task complexity score} (TCS) is, which follows from substitution from lemma~\ref{lemma:repr_alignment} and \ref{lemma:complexity}.
\end{corollary}

From the above calculations, we distill that the estimated coding length $\hat{H}_q(x)$ can potentially provide insights on the task complexity and confidence of task PAC-bounds. 
The derived bounds are not tight since, among other assumptions, many constraints within the weights are not taken into account by assuming independent log-normally distributed weights.
Nevertheless we observe experimentally that the quantitative differences in variance of weights as well as empirical entropy have a notable effect on training outcomes.

\section{Experiments}
The theoretical findings we presented in the previous sections have to be substantiated in the real world. We therefore conducted a wide ranging empirical study on their relevance over a number of different datasets and network architectures. In total we evaluated over 8000 networks of varying sizes.
We provide the full code and data required to replicate our experiments at the following URL:\\
\url{https://drive.google.com/open?id=1D8wICzJVPJRUWB9y5WgceslXZfurY34g}

\subsection{Datasets}
To have a chance that our findings are applicable beyond the scope of this publication we chose a diverse set of three datasets that capture different vision tasks. 
We chose two datasets for classification (KDEF and Groceries) and one for regression (Drone Racing). Sample images and further details for each of the datasets can be found in the appendix Tab.~\ref{tab:datasets}.

\textbf{KDEF:}
This dataset is based on an emotion recognition dataset by \citet{lundqvist1998karolinska}. 
Each of the images shows male and female actors expressing one of seven emotions. Images are captured from a number of different fixed viewpoints and centered on the face. To add more diversity to the data we added small color and brightness perturbations and a random crop. Moreover, since the dataset provides few samples, we downsampled each of the images to a sixth of their original size. \medskip \\
\textbf{Drone Racing:}
This dataset is based on the Drone Racing dataset by \citet{delmerico2019we}. We use the mDAVIS data from subsets 3, 5, 6, 9, and 10. While the original dataset provides the full pose (all six DOF), we train our feedforward networks to recover only the rotational DOF (roll, pitch, and yaw) from grayscale images. We matched the IMU data, which is sampled at 1000Hz to the timestamp for each grayscale image captured at 50Hz using linear interpolation. Since the images do not have multiple color channels we did not investigate YCbCr or PREC representations for this dataset. \medskip \\
\textbf{Groceries:}
We use the Freiburg Groceries Dataset~\citep{jund2016freiburg} and their original ``test0/train0'' split. Our only modifications are that we reserve a random subset of the test data for the evaluation of our hyperparameter optimization and that we reduce the size of the images from $256\times256$ to $120\times120$ pixels. Each of the images has to be classified into one of 25 categories.

\subsection{\emph{TCS} and Empirical Entropy Estimation}
To estimate the \emph{TCS}, we run ridge regression (RR) in batches of $256$ images with a small regularization multiplier $\lambda=0.1$ on different representations and feature extractions of the data. To approximate fully connected networks, we run RR on the vectorized images. To approximate convolutional networks we first convert each image to tiles of $7\times 7$ pixels with stride $2$ and then vectorize. We record the variance of learned weights across $10$ runs. 
To compute $\hat{H}_\mathcal{N}(x)$, we assume a multivariate Gaussian distribution on the variables and compute $\hat{H}_\mathcal(x)= \frac{1}{2} \log(\det(2\pi e \Sigma))$, where $\Sigma$ denotes the covariance matrix of $x$ and the respective feature extractions defined above are used. To calculate this we apply an SVD decomposition and use the sum of log singular values to estimate the entropy.

\subsection{Bayesian Hyperparameter Optimization}
Manually tuning hyperparameters for neural networks is both time consuming and may introduce unwanted bias into experiments. There is a wide range of automated methods available to mitigate these flaws~\citep{bergstra2012random, Hutter2015}.
We utilize Bayesian optimization to find the set of suitable hyperparameters for each network. 
Since the initial weights of our networks are sampled from a uniform random distribution we can expect the performance to fluctuate between runs. Due to its probabilistic approach Bayesian optimization can account for this uncertainty~\citep{shahriari2015taking}. 
The dimensions and their constraints were chosen to be identical for each representation but were adapted to each dataset. For an in-depth introduction to Bayesian optimization we refer to \citet{NIPS2012_4522}. More information on our particular implementation of Bayesian optimization can be found in the appendix.

\subsection{Network Architectures}

We investigated three basic architectures. Our optimization was constrained to the same domain for all representations of a dataset for each of the network architectures. 
The full list of constraints for each network and dataset can be found in the 
code accompanying this paper.
The initial learning rate was optimized for all architectures.

\textbf{Convolutional Networks:}
We use a variable number of convolutional layers~\citep{lecun1998gradient}, with or without maxpooling layers between them, followed by a variable number of fully connected layers. Moreover, kernel sizes and number of filters are also parametrized. \medskip \\
\textbf{Dense Neural Networks:}
These networks consist of blocks of variably sized fully connected layers. We optimize the activation function after each layer as a categorical variable. \medskip \\
\textbf{ResNets:}
The Residual Neural Networks or ResNets~\citep{DBLP:journals/corr/HeZRS15} are made up of a variable number of residual layers. Each of the layers contains a variable number of convolutions which themselves are fully parametrized.

\subsection{Representations}
While one could select an arbitrary number of representations for images, we limit ourselves to five which have previously been used in image processing.
Our focus is not on findings the best representations but to show how sensitive learning processes are to the representation of input data. 

\textbf{RGB:}
RGB is likely the representation we use most in our everyday life. Almost all modern displays and cameras capture or display data as an overlay of red, green and blue color channels. For simplicity we refer to the grayscale images of the Drone Racing dataset as being ``RGB''. \medskip \\
\textbf{YCbCr:} The YCbCr representation is used in a number of different image storage formats such as JPEG and MPEG~\citep{jpeg2000standard}. It represents the image as a combination of a luminance and two chrominance channels. It is useful for compression because the human eye is much less sensitive to changes in chrominance than it is to changes in luminance.\medskip \\
\textbf{PREC:}
This representation partially decorrelates the color channels of the image based on previous work on preconditioning convolutional neural networks~\citep{liu2018rotate}. For an image $\data \in \realnumbers^{m \times n \times c}$ with $c$ channels we first calculate the expected value of the covariance between the channels for each image in the dataset:
$
    \Sigma = \expected_{\dataset}\expected_{\data_{ij}} [ \data_{ij} \data_{ij}^T ] \in \realnumbers^{c\times c}$,
where $\data_{ij} \in \realnumbers^{c}$ is the channel vector at pixel $(i,j)$ of image $\data \in \dataset$.
We then solve the eigenvalue problem for $\Sigma$ obtaining real eigenvalues $\eigval$ and $\eigvec$ containing the eigenvectors. A small $\epsilon$ is added for numerical stability. u is stored in memory and consecutively applied to each image in the dataset. We get
$    U = \text{diag}(\eigval+\epsilon I )^{-\frac{1}{2}} \eigvec$
    which yields
    $\data_{\text{prec}} = \data_{\text{rgb}} \convolve U$. \medskip \\
\textbf{DCT:}
The 2D type II discrete cosine transform (DCT) is a frequency-based representation. Low frequency coefficients are located in the top left corner of the representation and horizontal/vertical frequencies increase towards the right or down, respectively. 
This representation applies the DCT transform to each of the channels separately. DCT has been used extensively for face detection~\citep{Hafed2001, 861296} and all its coefficients bar one are invariant to uniform changes in brightness~\citep{1427770}. \medskip \\
\textbf{Block DCT:}
Unlike for the DCT representation we apply the discrete cosine transform to $8\times8$ non-overlapping patches of each of the channels. This exact type of DCT is widely used in JPEG compression by quantizing the coefficients of the DCT and applying Huffman encoding~\citep{jpeg2000standard}.

\subsection{Training}
Each network was trained using an Adam optimizer~\citep{kingma2014adam}. Training was terminated when there were more than 7 previous epochs without a decrease in loss on the validation set or after 30 epochs. 

\section{Discussion and Results}

\begin{figure}[ht]
     \centering
    \begin{subfigure}[b]{0.75\textwidth}
         \centering
         \includegraphics[width=\textwidth]{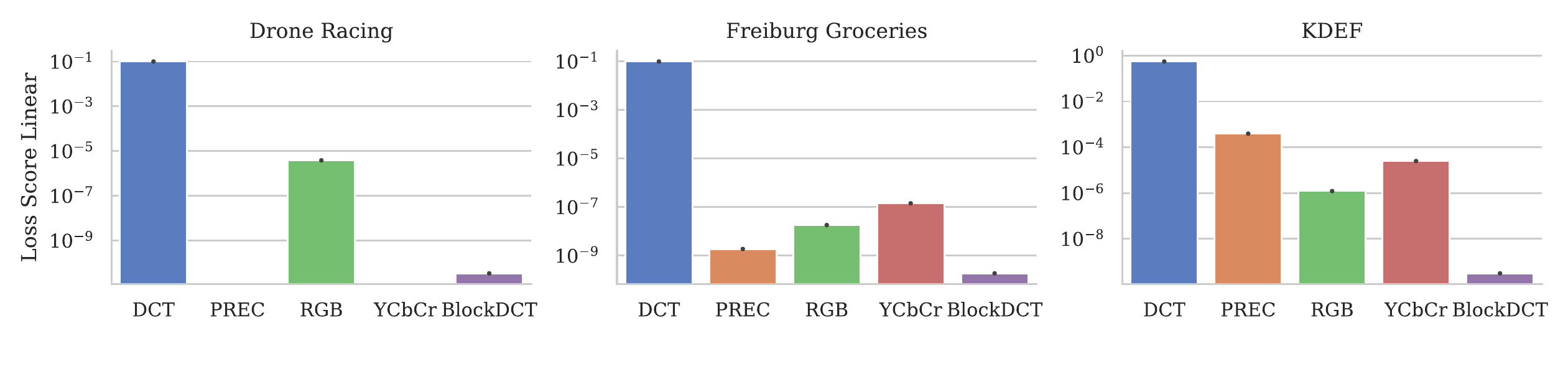}
     \end{subfigure}
     \hfill
    \begin{subfigure}[b]{0.75\textwidth}
         \centering
         \includegraphics[width=\textwidth]{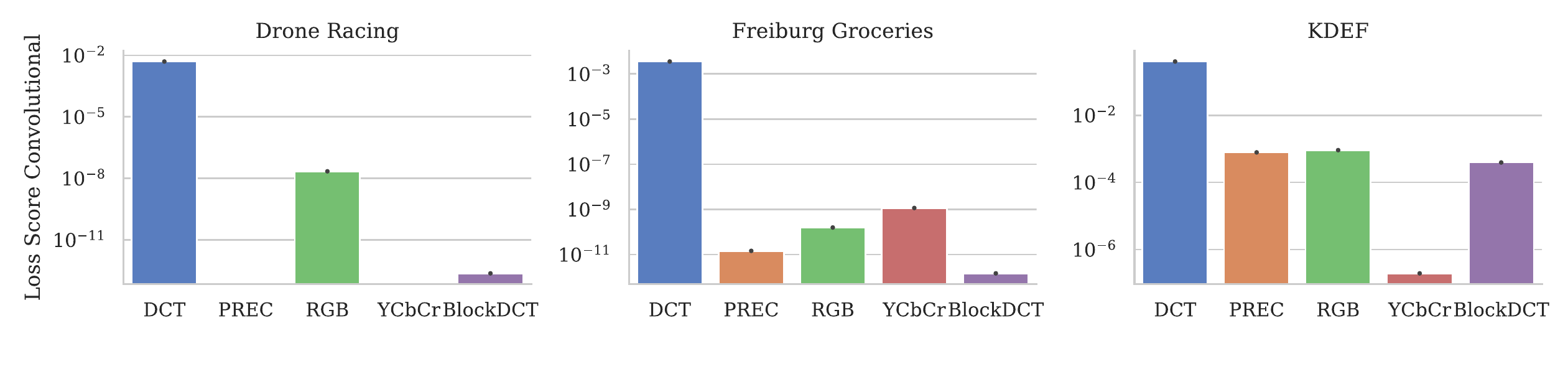}
     \end{subfigure}
    \caption{Linear and convolutional average \emph{batch training loss} scores for all datasets and representations. In particular this highlights the difficulty predicting the output from the DCT representation.}
    \label{fig:loss_rep}
\end{figure}
\begin{figure}[ht]
     \centering
      \begin{subfigure}[b]{0.75\textwidth}
         \centering
         \includegraphics[width=\textwidth]{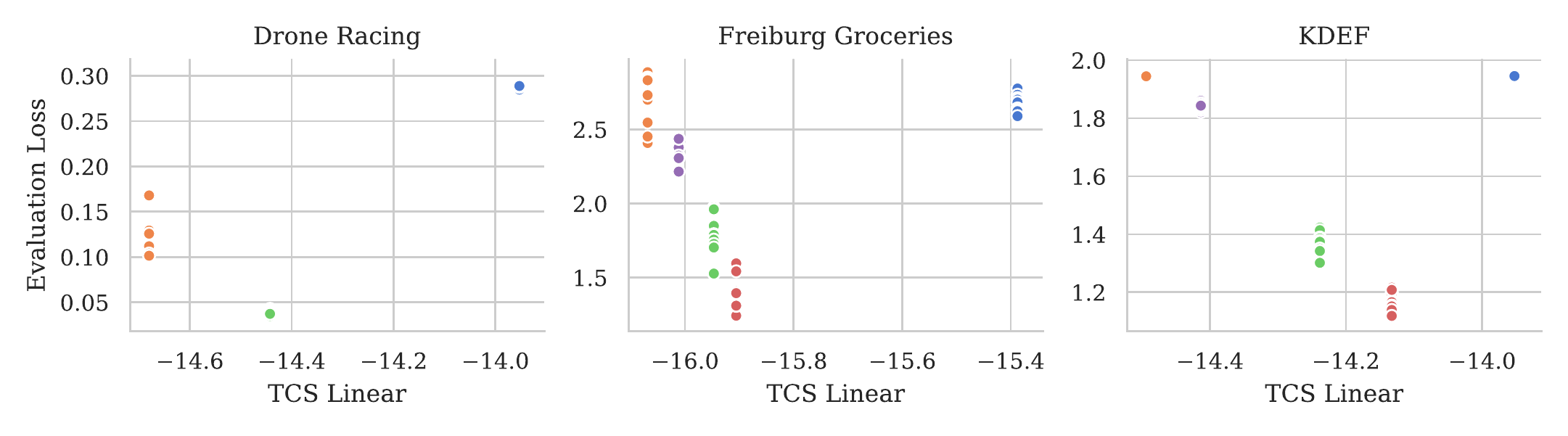}
     \end{subfigure}
     \hfill
     \begin{subfigure}[b]{0.75\textwidth}
         \centering
         \includegraphics[width=\textwidth]{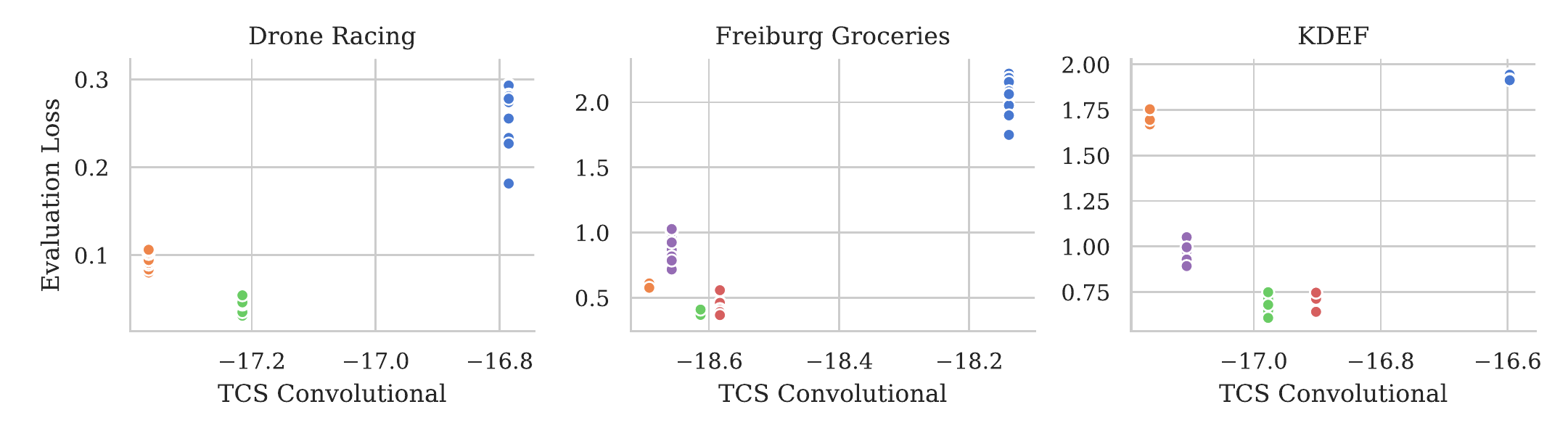}
     \end{subfigure}
    \caption{TCS values (shown in nats) correlate with the 10 best evaluation losses for each dataset and representations of RGB, YCbCr, Block DCT, PREC, DCT (green, red, orange, violet, blue). Scores associated to the DCT representation are disregarded due to insufficient training performance. Better scores relate to a higher confidence in predictions which are only worthwhile if training performance is satisfactory.}
    \label{fig:tcs_scores}
\end{figure}
\begin{figure}[ht]
    \centering
    \includegraphics[width=\textwidth]{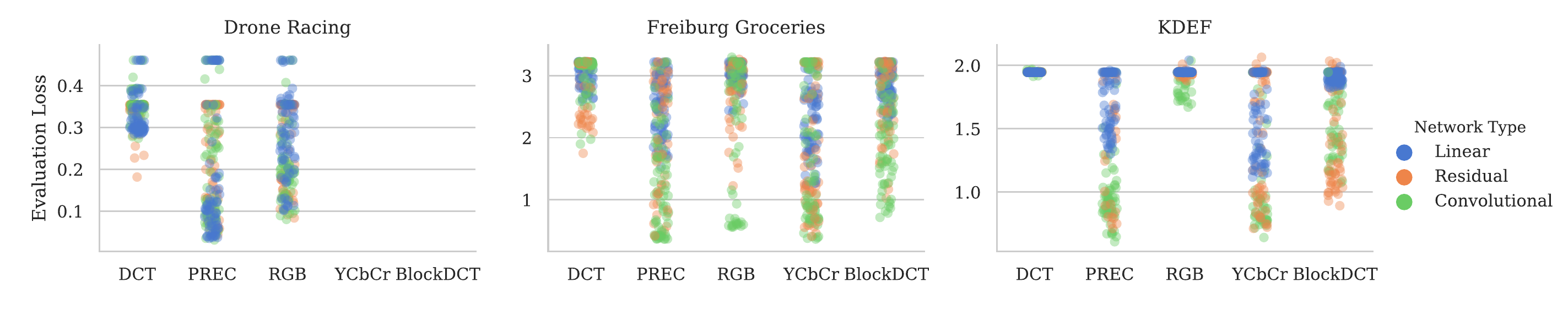}
    \caption{Performance of representations on the datasets for different neural network types.}
    \label{fig:raw_stats_plot}
\end{figure}

After evaluating a total of 5753 networks, 3702 of which finished training, we have verified our intuition that representations are important. We see a fairly consistent pattern over all datasets of RGB and YCbCr being the best representations, followed by PREC and blockwise DCT, while DCT falls short (see Fig.~\ref{fig:raw_stats_plot}). 
Moreover, we observe the great importance of hyperparameters in the spread of results for each network architecture. Had we chosen to hand-tune our parameters and accidentally picked a very poor performing network for the RGB representation we could have possibly come to the conclusion that DCT achieves better results on all datasets. 
As predicted, the performance of a representations also depends greatly on the network architecture. 

The \emph{TCS} scores we proposed show strong correlation with the results we obtained from architecture search if previous \emph{training batch loss} is used as a first filter criterion.
They can even predict the comparatively small differences between the other representations with reasonable accuracy (see Fig.~\ref{fig:tcs_scores}). 
Overall, we observe the significant correlated effect representation has on \emph{TCS} scores, estimated entropy (see Tab.~\ref{tab:results_table}) as well as performance.

\section{Conclusion}
This work started by trying to evaluate the effect different representations can have on task complexity. 
To achieve this we utilized a task complexity heuristics \emph{TCS}, as a score that takes both desired training performance and confidence of such an estimate into account.
Relevant to the confidence term of the task complexity, we derived a relation connecting mutual information of weights and dataset with the representation dependent expected coding length. 
As outlined, making full use of this perspective will depend to a great deal on how well we understand the candidate distributions of network architectures and representations that are currently in use. 
If the score proves itself also in future works, it may serve as a useful tool for automatic representation or architecture search.

\newpage
\bibliography{itml2019_bib}
\newpage

\section*{Appendix}
\subsection*{Results Table}

\begin{table}[h!]
% \begin{center}
\centering
\caption{Dataset overview}
\label{tab:datasets}
\begin{tabular}{lrrrr} 
\toprule
Dataset & input & output & split (train/val/test) & loss\\ 
\midrule 
Drone & $346\times260\times1$ & $3\times1$ & 10806/1403/1339 & L1\\
Groceries & $120\times120\times3$ & $25\times1$ & 6409/1018/1016 & NLL \\
KDEF & $91\times127\times3$ & $7\times1$ & 3931/470/497 & NLL\\ 
\bottomrule
\end{tabular}
% \end{center}
\end{table}

\begin{table}[ht]
\caption{Estimated linear (lin) and convolutional (conv) average training loss on batches of size 256, natural logarithm of TCS values and input data entropy assuming a normal distribution $H_\mathcal{N}(X)$ for all representations and datasets.}
\label{tab:results_table}
\begin{tabular}{@{}llrrrrrr@{}}
\toprule
                           &                & \multicolumn{3}{c}{linear} & \multicolumn{3}{c}{convolutional} \\ \cmidrule(l){3-8} 
Dataset                    & Representation & Train loss       & \emph{TCS}      & \emph{$H_\mathcal{N}(X)$}   & Train loss         & \emph{TCS}         & \emph{$H_\mathcal{N}(X)$}     \\ \midrule
\multirow{5}{*}{Groceries} & Block DCT      & 1.9e-10   & -16.07   & 443.7 & 1.5e-12    & -18.69      & 598.5   \\
                           & PREC           & 1.9e-09  & -16.01   & 376.7 & 1.5e-11    & -18.65      & 531.9   \\
                           & DCT            & 9.9e-02  & -15.39 & -187.9 & 3.6e-03    & -18.13    & -61.43   \\
                           & RGB            & 1.8e-08  & -15.95   & 346.7 & 1.6e-10    & -18.61       & 501.1   \\
                           & YCbCr          & 1.4e-07   & -15.91   & 282.3 & 1.2e-09     & -18.58      & 437.1   \\ \midrule
\multirow{3}{*}{Drone}     & Block DCT      & 3.4e-11  & -14.68   & 480.5 & 2.3e-13    & -17.37       & 638.7   \\
                           & DCT            & 1.0e-01  & -13.95  & -220.4 & 5.0e-03    & -16.79     & -79.34   \\
                           & RGB            & 3.9e-06  & -14.44    & 344.0 & 2.2e-08    & -17.22      & 500.86   \\ \midrule
\multirow{5}{*}{KDEF}      & Block DCT      & 3.1e-10  & -14.49   & 422.4 & 4.0e-04    & -17.16      & 576.8   \\
                           & PREC           & 3.9e-04   & -14.41   & 341.3 & 7.8e-04    & -17.10      & 496.4   \\
                           & DCT            & 5.5e-01  & -13.95 & -194.9 & 4.0e-01     & -16.60    & -189.2   \\
                           & RGB            & 1.2e-06    & -14.24   & 225.3 & 9.1e-04    & -16.98      & 381.4   \\
                           & YCbCr          & 2.5e-05   & -14.13   & 152.3 & 2.0e-07     & -16.90      & 307.6   \\ \bottomrule
\end{tabular}
\end{table}

To narrate Tab.~\ref{tab:results_table} we note that that estimating small entropy values is very error prone. 
When assuming a normal distribution, the entropy is calculated via the sum of the logarithm of eigenvalues of the covariance matrix of the data. 
The conditioning of the logarithm however gets worse, the closer its argument is to zero. Eigenvalues close enough to zero are thus likely to carry a significant error when used for entropy computation which is particularly prevalent in the DCT representation. 

\subsection*{Bayesian Optimization Supplementary Information}
\label{subsec:bayessupplement}
While there have been some theoretical proposals that would allow Bayesian optimization to be run in parallel asynchronously~\citep{NIPS2012_4522}, we restrict ourselves to a simple form of batch parallelization evaluating $n=6$ points in parallel. We acquire the points by using the minimum constant liar strategy~\citep{chevalier:hal-00732512}. The base estimator is first used after 10 points have been evaluated. 
Our acquisition function is chosen at each iteration from a portfolio of acquisition functions using a GP-Hedge strategy, as proposed in~\citep{DBLP:journals/corr/abs-1009-5419}. We optimize the acquisition function by sampling it at $n$ points for categorical dimensions and 20 iterations of L-BFGS~\citep{Liu1989OnTL} for continuous dimensions.

Since the optimizer has no information about the geometric properties of the network or if the network can fit in the systems memory, some of the generated networks cannot be trained. Two common modes of failure were too many pooling layers (resulting in a layer size smaller than the kernel of subsequent layers) and running out of memory, which was especially prevalent for dense networks. In our experiments we observed that roughly 35\% of all networks did not complete training. To stop the Bayesian optimizer from evaluating these points again we reported a large artificially generated loss to the optimizer at the point where the network crashed. The magnitude of this loss was chosen manually for each dataset to be roughly one order of magnitude larger than the expected loss. The influence of this practice will have to be investigated in future research.

\pagebreak
\subsection*{Representation Samples}

\begin{figure}[ht]
     \centering
     \begin{subfigure}[b]{0.18\textwidth}
         \centering
         \includegraphics[width=\textwidth]{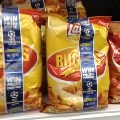}
     \end{subfigure}
     \hfill
     \begin{subfigure}[b]{0.18\textwidth}
         \centering
         \includegraphics[width=\textwidth]{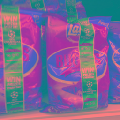}
     \end{subfigure}
     \hfill
     \begin{subfigure}[b]{0.18\textwidth}
         \centering
         \includegraphics[width=\textwidth]{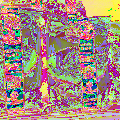}
     \end{subfigure}
     \hfill
     \begin{subfigure}[b]{0.18\textwidth}
         \centering
         \includegraphics[width=\textwidth]{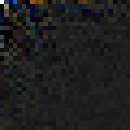}
     \end{subfigure}
     \hfill
     \begin{subfigure}[b]{0.18\textwidth}
         \centering
         \includegraphics[width=\textwidth]{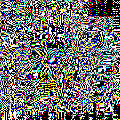}
     \end{subfigure}
        \caption{Image from the Groceries dataset in various representations. From left to right: RGB, YCBCR, PREC, DCT, blockwise DCT (cropped to show relevant coefficients, contrast boosted).}
        \label{fig:rep_overview}
\end{figure}

\subsection*{Dataset Samples}

\begin{figure}[ht]
    \begin{center}
    \resizebox{\textwidth}{!}{%
    \includegraphics[height=3cm]{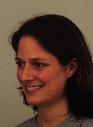}%
    \quad
    \includegraphics[height=3cm]{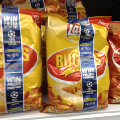}%
    \quad
    \includegraphics[height=3cm]{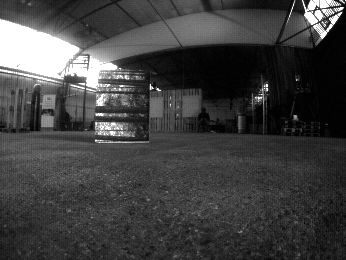}%
    }
    \end{center}
    \caption{Samples from each of the datasets. From left to right: KDEF, Groceries and Drone Racing.}
\end{figure}

\end{document}